# Depth Not Needed - An Evaluation of RGB-D Feature Encodings for Off-Road Scene Understanding by Convolutional Neural Network

*Christopher J. Holder*[1,2]     *Toby P. Breckon*[1]     *Xiong Wei*[2]

[1]School of Engineering & Computing Sciences, Durham University, UK
[2]Institute for Infocomm Research, Singapore

**ABSTRACT**

Scene understanding for autonomous vehicles is a challenging computer vision task, with recent advances in convolutional neural networks (CNNs) achieving results that notably surpass prior traditional feature driven approaches. However, limited work investigates the application of such methods either within the highly unstructured off-road environment or to RGBD input data. In this work, we take an existing CNN architecture designed to perform semantic segmentation of RGB images of urban road scenes, then adapt and retrain it to perform the same task with multi-channel RGBD images obtained under a range of challenging off-road conditions. We compare two different stereo matching algorithms and five different methods of encoding depth information, including disparity, local normal orientation and HHA (horizontal disparity, height above ground plane, angle with gravity), to create a total of ten experimental variations of our dataset, each of which is used to train and test a CNN so that classification performance can be evaluated against a CNN trained using standard RGB input.

## 1. INTRODUCTION

Scene understanding is a widely researched topic, and although the majority of scene understanding work only considers conventional 2D colour images (RGB), it has been demonstrated that the addition of 3D depth information (D) can improve classification performance in semantic scene understanding tasks, whether a laser scanner [1], stereoscopic camera [2] or structured light sensor [3] is utilised to provide a combined colour-depth (RGB-D) input.

More recently, deep-learning based methods have come to dominate scene understanding research, with convolutional neural networks (CNNs) achieving state of the art results on many image recognition tasks [4] [5]. The approaches of Fully Convolutional Network (FCN) [6] and SegNet [7] demonstrate leading performance for full pixelwise labelling of RGB images in a wide variety of contexts, particularly with SegNet in road scene understanding.

Prior attempts to utilise depth information in CNN classification tasks have shown that such features can bring about an improvement over results obtained from standard colour images: Gupta et al. [8] encode each pixel height, orientation and disparity into a 3 channel depth image to achieve state-of-the-art RGB-D object detection results, while Pavel et al. [9] utilise Histogram of Oriented Gradients and Histogram of Oriented Depth, computed from the output of a consumer depth camera (Microsoft Kinect), to obtain competitive results in an object segmentation task.

There is, however, relatively little scene understanding work focused on the off-road environment, where distinct but visually similar classes and a lack of regular structure can pose particular problems for computer vision. The approach of Jansen et al. [10] classifies pixels using colour based features and Gaussian Mixture Models, while Manduchi et al. [1] uses a combination of features from colour imagery and 3D geometry from a laser range-finder. In [11] a CNN is used to classify off-road imagery, with promising results achieved using only scene colour.

In this work, we examine the relevance of depth information within this context. Following the work of [11], we adapt the SegNet CNN architecture [7] to take multi-channel images comprising three RGB channels plus some combination of depth features ($D_{f1}$ $D_{f2}$ ... $D_{fn}$) as input. Using our own stereoscopic off-road data set, we create ten different encodings and train a network to perform pixelwise classification on each.

## 2. METHODOLOGY

Our method comprises a CNN architecture that takes multiple-channel input images and outputs a single class label for every pixel. Input images consist of 3-channel RGB information combined with one or more additional channels containing some 3D information derived from stereo disparity.

### 2.1. Network Architecture

The convolutional neural network architecture we use is nearly identical to that of SegNet, as described in [7]. We carry out minor alterations so that it can take multiple-channel images as input, and adapt the final layer to output the six classes found in our off-road scene data set.

The SegNet architecture, is comprised of a symmetrical network of thirteen 'encoder' layers followed by thirteen 'decoder' layers. The encoder layers correspond to the convolution and pooling layers of the VGG16 [12] object classification network, while the decoder layers up-sample their input so that the final output from the network has the same dimensions as the input. During the encoding phase, each pooling layer down-samples its input by a factor of two and stores the location of the maximum value from each 2 x 2 pooling window. During the decoding phase, these locations are used by the corresponding up-sampling layer to populate a sparse feature map, with the convolution layers on the decoder side trained to fill the gaps. This technique facilitates full pixel-wise classification in real-time, making SegNet an ideal architecture for use in autonomous vehicle applications.

For each of our input configurations, we initialise a network with randomised weights and train for 20,000 iterations, an amount empirically found to give the network ample time to achieve optimum performance. We use stochastic gradient descent (backpropagation) with a fixed learning rate of 0.001 and a momentum of 0.9, the values originally used to benchmark the SegNet architecture.

## 2.2. Data

We train the network using our own off-road data set, using images captured by stereo camera at an off-road driving centre in England. Our images have a resolution of 480x360 with a stereo baseline of 400mm giving us good depth resolution for a considerable distance in front of the vehicle. We manually label our ground truth images, with each pixel being assigned one of six class labels: {*sky, water, dirt, grass, bush, tree*}. Ten variants of the data set are created, each with one of five combinations of RGB-D features and one of two stereo algorithms, and each set is split into 80% training and 20% testing data sets.

## 2.3. Stereo Disparity

We compute two sets of stereo disparities, D, using Semi Global Block Matching (SGBM) [13] and Adaptive Support Weights (ASW) [14] methods, from which we also compute height above ground plane (H), normal orientation (N) and angle with gravity (A) of each point.

Stereo disparity measures the distance, in pixels, between the horizontal position of a feature in the left and right images of a rectified stereo image pair. Our data set contains disparities of between 1 and 64 pixels, which translates to a theoretical range of between 3 metres and 200 metres in front of the camera.

SGBM is widely used in real-time applications due to its combination of accuracy and speed. The algorithm first calculates the cost for every potential disparity match, in our case using the method from [17], and then attempts to minimise a global smoothness constraint while aggregating these costs along a one-dimensional path across the image.

In the ASW method, a local support weight window is computed for each pixel taking account of colour as well as spatial proximity, so that when matching is performed, neighbouring pixels that are part of the same surface, and therefore likely to be similar in colour to the pixel being matched, are given precedence. Despite being a more computationally complex algorithm, ASW has been shown to be easily parallelisable [18], allowing for real-time performance when run on a GPU. Furthermore, ASW has been empirically shown to provide a greater level of depth texture granularity than SGBM due to its ability to account for surface edges and the absence of a global smoothness prior.

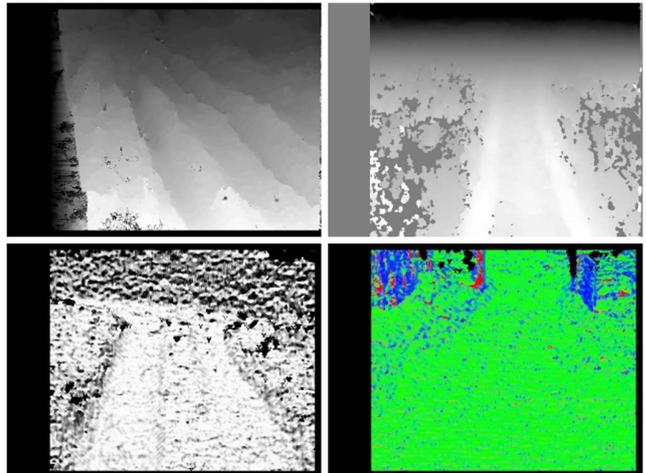

*Fig. 1 Four of the methods we use to encode depth data (clockwise from top left): disparity computed using the ASW method; Height above an assumed ground plane; Surface Normal (R, G and B values represent X, Y and Z respectively); Angle with gravity.*

For the disparity channel included in the CNN input, we normalise each stereo depth output to 0-255 to create an 8-bit image.

From this disparity information, we derive three further four-channel encodings: RGBD, RGBH and RGBA; and two six-channel encodings: RGBN, with *Nx, Ny,* and *Nz* each comprising one channel, and RGBDHA, combining disparity, height above ground plane and angle with gravity in the same manner as HHA encoding in [8]. These five encoding types combined with two stereo algorithms give us a total of ten data sets, in addition to our original RGB data set, for evaluation. Fig 1 visualises four of our depth encoding methods.

## 2.4. Height Above Ground Plane

From each disparity map we create height map H, encoding each pixel height above an assumed ground plane. Due to the rough nature of off-road environments, fitting a ground plane, as in [8] is not reliable, so we assume a ground plane fixed relative to the camera based on the known vehicle dimensions.

Each pixel height above this plane can be calculated using the pixel disparity value and known camera parameters. We clamp this value between 0 and 2*h*, where *h* is the height of the camera above the ground, then normalise to 0-255 to create an 8-bit image.

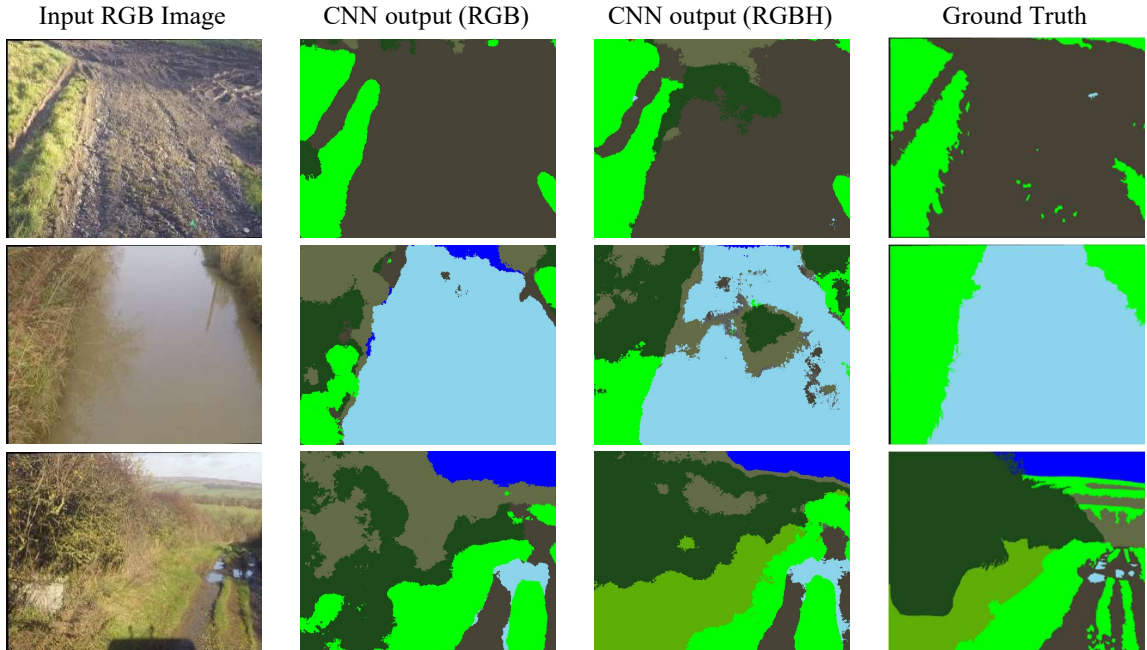

*Fig. 2 CNN output labels generated from RGB input, compared to those generated with the addition of a channel encoding pixel height above ground plane (RGBH), computed using semi global block matching disparities.*

### 2.5. Normal Orientation

A normal map $N$ encodes local orientation information at each point in an image. By calculating the 3D positions of three points surrounding each pixel, we can describe an imagined plane encompassing them, the orientation of which gives us a local normal vector. We multiply each component of this vector by 255 to create an image of three 8-bit channels encoding the normal $X$, $Y$ and $Z$ at each pixel.

### 2.6. Angle with Gravity

In [8], angle with gravity was found to be a useful feature when performing RGB-D object classification with a CNN. We compute each pixel angle with gravity, $A$, by comparing its normal vector with a gravity vector as in equation 1. As our data does not include vehicle orientation information and plane fitting is unreliable in the off-road environment, we assume gravity to always be (0, -1, 0) relative to the camera. Each pixel value is then multiplied by 255 to create an 8-bit image.

$$A_{x,y} = N_{x,y} \cdot (0, -1, 0) \quad (1)$$

## 3. EVALUATION

We assess our trained CNNs using the testing portion of our data set. Our primary metric for assessing classification performance is overall accuracy, defined as the number of correctly labelled pixels divided by the total number of labelled pixels in the test data. We also observe the mean average precision and recall over our six classes.

Table 1 shows the performance of each variant of input.

The best performing configuration was RGBH (height above ground plane) generated from SGBM stereo disparity, achieving an overall accuracy of 0.88. This shows that height is a particularly useful feature when considering the problem of off-road classification, where several classes can be delineated fairly consistently by their vertical positions.

Overall results obtained from ASW stereo were not as good as those from SGBM, which would seem to suggest that the additional encoded detail is of limited utility, and cannot counter the adverse effects of the extra noise introduced by the lack of any smoothness constraint. The worst performing encoding with both stereo algorithms was RGBN, which re-encodes a single disparity channel as three normal orientation channels. This may indicate that the addition of extra data channels can impact the ability of the CNN to converge on an optimal configuration, as a channel containing information that is redundant or otherwise not useful effectively decreases the signal to noise ratio of the input data. Despite this, the six channel RGBDHA configuration derived from SGBM stereo, performed fairly well, with a mean average precision higher than that of the RGB benchmark, suggesting that by careful selection of the data contained within additional channels, they can prove useful, even when they contain different encodings of identical information.

| Input format | Overall Accuracy | Mean Average Precision | Mean Average Recall |
|---|---|---|---|
| RGB | 0.87 | 0.82 | 0.85 |
| RGBD (SGBM) | 0.86 | 0.76 | 0.85 |
| RGBA (SGBM) | 0.84 | 0.81 | 0.78 |
| RGBH (SGBM) | **0.88** | 0.84 | **0.87** |
| RGBN (SGBM) | 0.83 | 0.8 | 0.77 |
| RGBDHA (SGBM) | 0.87 | 0.86 | 0.84 |
| RGBD (ASW) | 0.87 | 0.81 | 0.85 |
| RGBA (ASW) | 0.87 | **0.88** | 0.83 |
| RGBH (ASW) | 0.84 | 0.82 | 0.83 |
| RGBN (ASW) | 0.82 | 0.8 | 0.8 |
| RGBDHA (ASW) | 0.84 | 0.84 | 0.79 |

*Table 1 Classification results from each of our input encodings, with best results underlined.*

Fig 2 demonstrates our CNN output, with results obtained from RGB images compared to those from SGBM RGBH images. A slightly greater amount of detail can be seen in the RGBH images, with edges appearing slightly better defined than in their RGB counterparts, however the noise present in the image containing a large body of water illustrates the inability of stereo techniques to cope with transparent or reflective surfaces, representing a key area for future work.

## 4. CONCLUSIONS

Our results show that the additional information contained in stereo disparity can provide marginal improvement to CNN classification performance compared to standard RGB images in this particularly challenging problem of off-road scene understanding. However, careful consideration of how this information is encoded is necessary, and in our case the combination of RGB data with each pixel's height above ground plane was shown to give the best results.

The performance gains demonstrated are small, however this is mostly due to the efficacy of the existing CNN architecture at classification from conventional colour images (RGB), with results so good to begin with that it becomes ever harder to gain extra performance.

This is counter to results such as those from Gupta et al [8] and Pavel et al [9], who claim that RGBD data provides significant improvements in CNN classification performance over RGB imagery. This demonstrates that while depth information may improve CNN performance in an indoor object detection task, it is of limited utility in the more challenging off-road environment, where poorly defined class boundaries and inconsistent object shape can hamper classification accuracy.